\documentstyle[11pt,fullpage]{article}

\newcommand{\ra}{\rightarrow}

\newcommand{\bottom}{\perp}

\title{The Compactness of Construction Grammars}
\author{Wlodek Zadrozny \\
IBM Research,
T. J. Watson Research Center \\
Yorktown Heights, NY  10598  \\
{\tt wlodz@watson.ibm.com} \
\thanks{IBM Research Report RC 20003 (88493)}}
\date{March, 1995}

\begin{document}
\maketitle

\begin{abstract}

We present an argument for {\em construction grammars}
based on the minimum description length (MDL) principle
(a formal version of the Ockham Razor). The argument consists in
using linguistic and computational evidence in setting up a
formal model, and then applying the MDL principle to prove
its superiority with respect to alternative models.
We show that construction-based representations are
at least an order of magnitude more compact that the corresponding
lexicalized representations of the same linguistic data.

The result is significant for our understanding of the relationship
between syntax and semantics, and consequently for choosing
NLP architectures. For instance, whether the
processing should proceed in a pipeline from syntax to semantics to
pragmatics, and whether all linguistic information should be
combined in a set of constraints. From a broader perspective,
this paper does not only argue for
a certain model of processing, but also provides a methodology for
determining advantages of different approaches to NLP.

\end{abstract}
%

\section{Introduction: Motivation and Terminology}

We present an argument for a particular model for natural language,
namely {\em construction grammars}
(cf.
\cite{Goldberg94} and
\cite{Fillmal88},
for a comprehensive introduction, and
\cite{Jurafsky92},
\cite{Ward92},
\cite{ppsith94},
\cite{Zad94sym},
\cite{cons1},
\cite{Coling94} for computational models
).
The result we report establishes their optimality, in the sense of
the minimum description length (MDL) principle
(a formal version of the Ockham Razor). The argument consists in
using linguistic and computational evidence in setting up a
formal model,
and then using the MDL principle to prove its superiority with
respect to alternative models.
The result is significant for our understanding of the relationship
between syntax and semantics, and consequently for choosing
NLP architectures. For instance, whether the
processing should proceed in a pipeline from syntax to semantics to
pragmatics, and whether all linguistic information should be
combined in a set of constraints. From a broader perspective,
this paper does not only argue for
a certain model of processing, but also provides a methodology for
determining advantages of different approaches to NLP.\\

After discussing the terminology,
we begin our exposition with a sketch of the argument (Section 1.2).
The argument will have two parts, of which the first is empirical
and based on the work of other researchers on language structure and
models of processing, and where the second part applies
the MDL principle. Both will be presented in Section 3.
For this second part we need
to develop some techniques and intuitions. Thus, in Section 2,
we present the details of the application of the MDL principle
for a grammar of numbers. Since the domain of numbers is completely
unambiguous, this exposition move will make transparent
the subsequent application of MDL principle to a grammar of English
in Section 3.2. We conclude the paper with a few open problems.

\subsection{Terminology}

By a {\em grammar} we mean a collection of entities, like
data structures or logical formulas, that describe a formal
language. Thus grammars are always formal. If the formal
language in question resembles English, we call the grammar a
{\em grammar of English} (ditto for other natural languages).

While it is possible to classify grammars along many dimensions,
in this paper we are interested in two: (1) Whether a grammar
is lexicalized, and (2) whether a grammar separates information
about form and meaning (syntax and semantics) or mixes the two.
Obviously, in each case these are mutually exclusive possibilities;
i.e. either all information about
language is contained in the lexicon or not, and the same for (2).
If a grammar is not lexicalized,
then it must refer to units larger than words, and we will call
such an item of information a {\em structural rule}; the most
typical example of which would be a phrase structure rule.

A {\em construction grammar} is a non-lexicalized grammar in which
information about form and meaning is kept together in constructions.
A {\em construction} is a set of constraints
about form and meaning of a word, sequence of words, or a sequence
of constructions. (A recursive definition).

This definition captures the idea that forms and meanings should
be investigated together. However we want to say more than that.
Since meaning very often depends on context, it is only natural to make
this connection explicit. Therefore in the formalism we use
constructions are triples:
$$ < Context; \  Form; \ Meaning >$$
The {\em Form} describes the construction as a combination of
subconstructions (e.g. a noun phrase as a combination of
a numeral and a noun); the {\em Meaning} \ part specifies how
the meanings of the subparts contribute to the meaning
of the construction (e.g. specifying the number of
of elements).
The {\em Context} \  specifies the parameters that are necessary to
construct the meaning, and which are not present in the meanings
of the parts; for example, the content of the question is necessary
to construct the meaning of an answer, especially if the answer
is just a sentence fragment.

\vspace*{.25in}
The {\em minimum description length (MDL) principle} was proposed by
Rissanen \cite{Rissanen82}. It states that the best theory to explain
a set of data is the one which minimizes the sum of
\begin{itemize}
\item the length, in bits, of the description of the theory, and
\item the length, in bits, of data when encoded with the help
of the theory.
\end{itemize}

In our case, the data is the language we want to describe, and the
the encoding theory is its grammar (which includes the lexicon).
The MDL principle justifies the intuition that
a more compact grammatical description is better.
At issue is what is the best encoding.
To address it, we will be simply comparing two complementary
classes of encodings and showing that one of them is usually more
compact. The formal side of the argument will be kept to
the minimum: after building the two complementary
models of language, the mathematics will
be simple --- counting. (But we will discuss
some ways of producing more refined models).

\subsection{The line of the argument}

The paper is about how to best represent information about language, i.e.
about data structures. As we know, data structures are
determined by both the
types of data and their structure. Therefore we have to argue
for particular types and particular structures. In our particular case,
to argue for grammars of constructions, we
will show that the data types should contain information
about both form and meaning. Secondly, we will show
that their structure should
contain something resembling phrase structure rules; we do it by
presenting an MDL-based argument against lexicalized
representation of forms and meanings.

The optimality argument goes as follows: we have to prove that
for NLP it is preferable not to separate syntactic, semantic
and pragmatic information --- which is an argument
for having data structures that combine them. But we could imagine
say lexicalized grammars in which such information is combined.
Hence,  as the second step, we have to show that grammars with
"phrase structure rules" are better than lexicalized grammars.
These two arguments show the superiority of a construction-based approach
with respect to alternative grammatical formalism.

\section{Using MDL with grammars of numbers}

To make the presentation clear we first discuss a grammar of
numbers --- for numbers our whole argument is completely
formal and transparent. In the next section we use the same argument
for NL grammars.\\

\noindent
{\bf A lexicalized grammar of numbers}

Any grammar of numbers must somehow express the fact
that the value of a digit $D$ depends on its position. I.e.
$$ \mu (D) = D * 10 ** pos(D) $$
where $pos(D)$ is the position of $D$ (counting from the right,
beginning with 0).
Notice that
a lexicalized grammar expresses it directly, and must repeat it for
every digit.
(We use the same symbols for the digits and their values).
\begin{description}
\item $ \ \  < [10]; \ 0 ; \ \mu (0) = 0  > $
\item $ \ \  < [10]; \ 1 ; \ \mu (1) = 1 * 10 ** pos(1)  > $
\item $ \ \  < [10]; \ 2 ; \ \mu (2) = 2 * 10 ** pos(2)  > $
\item $ \ \ \ \ \ \ \ \ \ \ ... $
\item $ \ \  < [10]; \ 9 ; \ \mu (9) = 9 * 10 ** pos(9)  > $
\end{description}

Based on that, the value of a number is  the sum of values of
its digits:
$$ \mu (N) = \sum_{D \ in \ N} \mu(D) $$

Note that the grammar specifies the formula for value of the type,
and the value of the token is given by its instantiation. E.g.
in computing
$\mu (17341)$, notice the different values of the digit $1$.\\
Also, note that given a different set of functions, e.g.
{\em head, tail, log, *, +}, the $\mu$ function would be slightly
different, but the lexicon must look essentially the same, because
there is no other place to put the data about how the forms determine
the meanings.\\

\noindent
{\bf A construction grammar of numbers}

The lexical part of a construction
grammar can now be much simpler. (And it could be
simplified even further by assuming that the value of a token
is the token, unless specified otherwise).
\begin{description}
\item $ \ \  < [10]; \ 0 ; \ \mu (0) = 0  > $
\item $ \ \  < [10]; \ 1 ; \ \mu (1) = 1   > $
\item $ \ \  < [10]; \ 2 ; \ \mu (2) = 2   > $
\item $ \ \ \ \ \ \ \ \ ... $
\item $ \ \  < [10]; \ 9 ; \ \mu (9) = 9  > $
\end{description}

In contrast to the lexicalized grammar,
in a grammar of constructions we can write a structural rule
$$ <[10]; \ DS \ra DS1 \ D \; \; ; \
 \; \mu(DS)=10*\mu(DS1) + \mu (D) > $$
This rule defines a production saying
that a new structure is obtained by adding a digit $D$ to
a previously defined structure $DS1$. The equality
associates the meaning of a new structure $\mu(DS)$ with
the meaning of its components $\mu(DS1)$ and $\mu(D)$.
As a consequence of choosing this kind of representation,
the rule about how to compute the meaning of digits
has to be stated only once.

As we can see, describing the same language with a grammar of
constructions results in a more compact grammar. We saved
11 symbols per non-0 lexical entry, i.e. 99 symbols altogether.
Although, we added a structural rule,
its size is comparable with the above
$\Sigma$ rule for computing the value of a sequence of digits
in the lexicalized
grammar. Also note that the latter must additionally refer to
the function $pos$ and exponentiation.

While savings 99 symbols is not much, for larger lexicons the saving
would be much bigger. Larger lexicons are obtained by increasing
the ${Base}$.
In this case the grammar production (phrase structure rule) reads:
$$ < [Base] ; \; DS \ra DS1 \ D \; \; ; \; \mu(DS) =
Base*\mu(DS1) + \mu (D) > $$
It can be easily checked that the resulting construction grammar
is always an order of magnitude more compact than its lexicalized
counterpart.

A grammar of constructions is even more compact if additional
conditions are placed on sequences of digits, e.g. that
only ascending sequences of digits are acceptable. That is so,
because typically any such a condition would have to be included
in all lexical entries, and the more complicated the condition,
the less compact is the lexicalized grammar.

\section{The superiority of constructions for NLU grammars}

In this section we present the argument that (a) it makes sense
to encode syntactic, semantic and pragmatic information together,
and (b) that construction-based
grammar are more compact that lexicalized grammar that encode
the same semantic information. At the end of the section we
discuss some possible extensions of the model to cover the case
of "lexicalized grammars with a few constructions".

\subsection{Data types: FORMS, MEANINGS,  and CONTEXTS}

There are three arguments supporting the encoding of linguistic
information in data structures that combine syntactic, semantic
and pragmatic information. Namely, the linguistic theory,
the practice of computational linguists who encode it that way,
and experimental evidence from analyzing parsing mechanisms.

We now briefly discuss each argument.
Regarding linguistic theory, the {\em Comprehensive Grammar of the
English Language}  \cite{Quirketal85} describes the language by freely
combining syntax, semantics and pragmatics. With a closer look,
the analysis of the structure of VPs and NPs requires the reference
to semantic information; e.g. McCawley
\cite{McCawley88}, vol.1, p.222 and ff.  argues
that the restriction on the progressive {\em be} is semantic rather than
syntactic in nature, that is,
its complement should refer to an {\bf activity}
or a {\bf process} rather than a {\bf state}.
Finally, \cite{Goldberg94} contains a comprehensive set of
arguments for construction-based description of clause structure.

Moving to relevant work in computational linguistics, first let us
notice that
in some cases the strategy of describing language structure using only
syntactic markers produces impressive results (e.g.
\cite{TapandV94}).
But when the goal is to understand language, with the increasing
coverage of a grammar the set of its markers
grows and encodes more and more semantics.
For instance PEG, a broad coverage grammar of English,
\cite{Jensen86},
\cite{PLNLP93},
used about 400 markers, including {\sc money,
date, phone, animate}, {\sc human, religious\_name}, {\sc time, title,
verb\_of\_cognition, verb\_of\_asking}, {\sc derogatory, emphatic} etc.
Furthermore,
if we examine computational linguistic literature we can see that
to prevent overgeneralizations,
semantic and pragmatic information must be taken into
account (see e.g.  \cite{Allen87},
\cite{HinkelmanandAllen89},
\cite{Hirst87}, \cite{JenseandBinot88}); and
it is virtually impossible to produce a correct
predicate-argument structure for many constructions (e.g.  PPs; relative
clauses; parallel, cumulative or periodic sentences)
without incorporating
those two kinds of information (see also
\cite{Fenstadetal87},
\cite{Ravin87}, \cite{Noonan85},
\cite{Keenan85}).
Clearly, the
integration of various types of information is necessary to interpret
discourse
(e.g. \cite{Hobal93}).

While the body of work we have quoted provides evidence from coverage and
depth of semantic processing, there is also evidence based on
the efficiency and robustness of parsing.
Thus, Lytinen \cite{Lytinen91}
shows that a semantically driven approach is
superior to syntax-first approach in processing text in narrow domains.  His
semantic-first algorithm decides which grammar rules to apply next on the
basis of "desirable semantic attachments between adjacent constituents", after
such constituents have already been identified.
Dowding et al. \cite{Dowdingetal94} show that the same strategy works
with a large coverage grammar of English.


\subsection{The structure of data:
Why the argument for numbers works also for NL}

To show that the argument of Section 2 works for a grammar
of English we should show that the use of structural rules
can result in a more compact grammar.

We need a simple model, so let us consider the problem  of
determining whether a clause followed by a set of PPs is nonsensical.
For example,\\
\hspace*{.4in}
{\em we meet at 12 with bob at 6 avenue and 44 street} vs. \\
\hspace*{.4in}
{\em the dow closed at 2200 with bob at 6 avenue and 44 street}.\\
In general, this problem cannot be solved without access to
a large body of background knowledge; so, let us simplify it further.
Assume that sentences that repeat the same type of information are
nonsensical, e.g.\\
\hspace*{.4in}
{\em we meet at 12 pm with bob from 5 to 6 pm} \\
To set up the model we have to define a formal language resembling
English. We do it in two steps. Let our first formal language $L_{1}$
consist of all sequences of SV (subject-verb),
SVO (subject-verb-object), SVOO (subject-verb-object-indirect object),
of English taken from some very large corpus. The language $L_{PP}$
we are interested in consists of sentences of $L_{1}$ followed by
any number of PPs (prepositional phrases) that contain the nouns
from $L_1$.
(Hence $L_{PP}$ is infinite).
Using this model we can discuss differences between
a construction grammar and a lexicalized grammar. \\

\noindent
{\bf Construction grammar} \\

\noindent
To define a construction grammar for $L_{PP}$ we define the
lexicon and a set of productions (phrase structure rules).
Let the lexical entries be given by the matrix:
$$ < []; \ w ; \ \mu(w) = \{ cat(type_i, w_i): \ i < n_w \}  > $$
i.e. the meaning of a word is given by its linguistic
category, a word
sense, and the semantic type of the word sense.
The meaning is given as a set, because words often belong to different
categories and may have multiple word senses,
e.g.
$$ \mu(dog)= \{ verb(pursue, dog_0), noun(person,dog_1),
                                   noun(person,dog_2),
                                   noun(animal,dog_3) \} $$

{\bf Remark.}
We ignore the context to make the argument more general. Also,
we could write $ < []; \ w ; \mu(w)=cat(w_i) > $, and associate
one word with multiple lexicon entries
(slightly abusing the notation).\\

Having defined the lexicon,
we have to cover the SV, SVO, and SVOO constructions; for instance
we could write the SVO-action  construction as
$$ < [] ; \;
CL \ra S \ V \ O \; \; ; \;
\mu(CL) = \left[ \left[action, \mu(V) \right],
\left[agent, \mu(S) \right],
\left[object, \mu(O) \right]  \right] > $$
However for our purposes it is irrelevant how the meanings of those
constructions are encoded; the only thing we will need is the
existence of the  $\mu$ function.

The next step is to cover the PPs. We represent them as
$$ < []; pp \ \ra \ prep(p) \ noun(type_i,w_i) ; \
\mu(pp) = pp(type_{i,p}, w_i) > $$
I.e. we assume that
prepositional phrases have types, and each type is
a function of the preposition and the noun type. E.g.
$$ < []; pp \ \ra \ prep(at) \ noun(hour,X) ; \
\mu(pp) = pp(event\_time,X) > $$
Thus "at" followed by the second sense of 2200 (as hour, not number),
would be classified as
an "event\_time". (We are conveniently assuming that
numerals are nouns).\\

Finally, we define the construction of adding an adjunct to a clause
(with + standing for the simple {\em append} or a more complex
procedure):
$$ < [] ; \
CL \ra CL1 \ A \; \ ; \;
\mu(CL) = \mu(CL1) + \mu (A) \;  \ \
\  {if}  \ \mu(A) \ not \ in \ \mu(CL1) \ , \ {and} \
\bottom - otherwise > $$
Thus defined construction grammar encodes our formal language
$L_{PP}$. \\

\noindent
{\bf Lexicalized grammar}

For a lexicalized grammar, as in Section 2, we observe that the meaning
of a clause such as the one above must be encoded with each noun.
Thus, for each noun we have to encode the meaning of its adjuncts, i.e.
repeat the formula:
\begin{quote}
"if I am combined with preposition $X_1$, then our type
will be $T=T_1$; \\
"if I am combined with preposition $X_2$, then our type
will be $T=T_2$; \\
...\\
and if $T$ does not appear in the meaning $E$ of everything
to the left \\
\ \ \ \ of the preposition immediately to the left
of me, then I make sense; \\
and our joint meaning of is the sum of $T$ and $E$."
\end{quote}
For instance, for each number that can denote an hour we would
have:

\[
\begin{array}{ll}
<  [] \ ; \ \ 2 \ ; & \ \ \mu (2) = \{(hour,2) \} \\
\ &  \  \  {if \ "at" \ then } \ \ T=event\_time \\
\ &  \  \  {if \ "from" \  then } \ \ T=beginning\_time \\
\ &  ... \\
\ &  \  \  {if \ T \  appears \ in \  E \
\  to \ the \ left \ } \ then \  \bottom  \\
\ &  \  \  \    \  {otherwise} \ add \ T \ to \  E \  > \\
\end{array}
\]
(In addition the formulas might encode
constraints on {\em event\_time},
{\em beginning\_time}, and other types).
As before, the reason for repeating this piece of information
is that these formulas must somehow be encoded.
Since the reference to anything larger than a word
is forbidden, the formula must be repeated for every word separately. \\

Repeating the argument of
Section 2,  we conclude that a construction grammar
that encodes the formal language $L_{PP}$ is at least an order
of magnitude more compact that any lexicalized grammar that
encodes this language.
The exact difference in bits can be computed as in
\cite{LiandVitanyi93} pp. 312-316, as a function of the number
of nouns in the language.

\subsection{Why adding a few structure rules won't help}

With any mathematical model of language one should ask the
question how closely it approximates the linguistic reality,
and what happens if we introduce some minor changes to it.
For example, what if a grammar is 99.99\% lexicalized and
contains only a few phrase structure rules (e.g. the
rule about the meanings of clauses with adjuncts).
The answer is that the same argument would apply, provided
we can find another productive grammatical phenomenon that
behaves similarly to digits in our grammar of numbers or to
PPs in the case of $L_{PP}$. Is there a reason to believe that
such productive phenomena are common? Yes.

The term to denote a situation where a language pattern is
productive, but has many exceptions, is
{\em partial productivity} (\cite{Goldberg94}).
One such pattern is the ditransitive construction ({\em ibid.}): \\
\hspace*{.25in}
{\em John faxed Bob the report.} \\
\hspace*{.25in}
{\em Barbara told Bob the story.}\\
\hspace*{.25in}
{\em *Barbara whispered Bob the story.}\\
To account for its partial productivity, a rule that describes its
usage must refer to the semantic types of the verb, the object,
and the indirect object. This situation is similar to the
one described in Section 3.2. Thus, the ditransitive construction
should not be lexicalized either.

Most linguistic phenomena exhibit this quality of partial productivity.
And to many the techniques of Section 3.2 could be applied.
These are for instance adjuncts, open idioms \cite{Fillmal88},
noun-noun modifications, the order of adjectives,
and long distance dependencies \cite{Jurafsky92}.
For all of them we could set up models similar to the ones
discussed above; furthermore, these models could be combined
in a bigger structure. Thus, by using a more comprehensive
model of English, we could have made the construction grammars
many orders of magnitude more compact than their lexicalized
counterparts.

This argument can be even further strengthened by looking
beyond the clause. Similar models  can be created for
connectives and linking words (such as "therefore", "consequently").
Moving to dialogs, where understanding sentence fragments requires
the context (e.g. about previous
discourse), we could build models, where the difference in
the size of
construction grammars and a lexicalized grammars would be due to
the necessity to encode the context (the simple task being e.g.
to identify reasonable and absurd answers). Using the techniques of
\cite{ZadandJen91} it is easy to show that the arguments
of Sections 2 and 3 could also be used with paragraph coherence,
and most likely extended beyond the paragraph.

\section{Open problems and Conclusions}

Since this paper is about counting, the most natural open
problem is the number of constructions beyond the word,
and then beyond the clause. Of course, it is not clear
how to count them. Still, we know approximately the size
of lexicons for spellchecking; similarly, we could ask about the size of
a construction grammar for NLU (e.g. at the level of
achieving a decent score on a SAT exam)
or for summarizing of New York Times stories.

The next question is whether using
techniques similar to those of Section 3.2
it can be shown that some particular ways
of representing constructions
are optimal (e.g. the embedding of ontology into the grammar or
the restriction to only two levels of structure, as in
\cite{Zad94sym}).

Finally, there are many questions that are not directly related
to the topic of this paper, e.g. the role of ontologies in
describing constructions, the interaction between structural
and functional descriptions etc.\\

{\bf Conclusions:}
We used a combination of empirical data
and the minimum description length principle
to show that construction grammars are
better representation of linguistic data than their
lexicalized counterparts.
This hybrid ---  linguistic, computational, and
mathematical --- argument for a construction-based
approach to language understanding is the main contribution of
this paper. However, it is likely similar arguments could be
used in other circumstances where one has to choose between
competing representations.

\bibliographystyle{plain}

\end{document}